%% file: main.tex
\documentclass{article}

\PassOptionsToPackage{numbers}{natbib}

\usepackage[preprint]{neurips_data_2022}

\usepackage[utf8]{inputenc} 
\usepackage[T1]{fontenc}    
\usepackage{hyperref}       
\usepackage{url}            
\usepackage{booktabs}       
\usepackage{amsfonts}       
\usepackage{nicefrac}       
\usepackage{microtype}      
\usepackage[dvipsnames]{xcolor}         

\newcommand{\answer}[1]{{\textcolor{BlueViolet}{#1}}}

\usepackage{algpseudocode}
\usepackage{algorithm}
\include{math}
\usepackage{caption}
\newsavebox{\tempfig}
\usepackage[pdftex]{graphicx}
\usepackage{subcaption}
\usepackage[many]{tcolorbox}
\usepackage{enumitem}

\newtheorem{definition}{Definition}

\title{FLAIR: Federated Learning Annotated Image Repository}

\author{%
  Congzheng Song\thanks{Equal contribution} \\
  Apple\\
  \texttt{csong4@apple.com}
  \And 
  Filip Granqvist$^*$ \\
   Apple \\
  \texttt{fgranqvist@apple.com} 
  \And
  Kunal Talwar \\
  Apple\\
  \texttt{ktalwar@apple.com}
}

\begin{document}

\maketitle

\begin{abstract}
Cross-device federated learning is an emerging machine learning (ML) paradigm where a large population of devices collectively train an ML model while the data remains on the devices.
This research field has a unique set of practical challenges, and to systematically make advances, new datasets curated to be compatible with this paradigm are needed.
Existing federated learning benchmarks in the image domain do not accurately capture the scale and heterogeneity of many real-world use cases. 
We introduce FLAIR, a challenging large-scale annotated image dataset for multi-label classification suitable for federated learning.
FLAIR has 429,078 images from  51,414  Flickr users and captures many of the intricacies typically encountered in federated learning, such as heterogeneous user data and a long-tailed label distribution.
We implement multiple baselines in different learning setups for different tasks on this dataset. 
We believe FLAIR can serve as a challenging benchmark for advancing the state-of-the art in federated learning.
Dataset access and the code for the benchmark are available at \url{https://github.com/apple/ml-flair}.

\end{abstract}

\input{introduction}
\input{related}
\input{prelim}
\input{data}
\input{experiments}
\input{discussion}

\input{conclusion}

\bibliographystyle{plain}
\bibliography{references}

\appendix
\input{appendix}

\end{document}

%% file: math.tex
\usepackage{amsmath,amsfonts,bm}

\def\eqref#1{equation~\ref{#1}}

\def\1{\bm{1}}

\DeclareMathAlphabet{\mathsfit}{\encodingdefault}{\sfdefault}{m}{sl}
\SetMathAlphabet{\mathsfit}{bold}{\encodingdefault}{\sfdefault}{bx}{n}

\def\gD{{\mathcal{D}}}

\def\gM{{\mathcal{M}}}

\def\gR{{\mathcal{R}}}

\newcommand{\prob}{\mathrm{Pr}}

\algrenewcommand\algorithmicindent{1.0em}%
\mathchardef\mhyphen="2D 



%% file: introduction.tex
\section{Introduction}

Remote devices connected to the internet, such as mobile phones, can capture data about their environment. Machine learning algorithms trained on such data can help improve user experience on these devices. However, it is often infeasible to upload this data to servers because of privacy, bandwidth, or other concerns.

Federated learning~\citep{mcmahan2017communication} has been proposed as an approach to collaboratively train a machine learning model with coordination by a central server while keeping all the training data on device. Coupled with differential privacy, it can allow learning of a model with strong privacy guarantees. Models trained via private federated learning have successfully improved existing on-device applications while preserving users' privacy~\citep{granqvist2020improving,hard2018federated,googleflblog}.

This has led to a lot of ongoing research on designing better algorithms for federated learning applications. Centralized (non-federated) machine learning has benefited tremendously from standardized datasets and benchmarks, such as Imagenet~\citep{imagenet}. To evaluate and accelerate progress in (private) federated learning research, the community needs similarly high quality large-scale datasets, with benchmarks. Ideally, the dataset would be representative of the challenges identified as important by the community~\citep{kairouz2021advances}. Additionally, the benchmark should provide common, agreed-upon metrics to allow comparison of privacy, utility, and efficiency of various approaches. 

Federated data may have various non-IID characteristics that are seldom encountered in traditional ML~\citep{kairouz2021advances}.
These include shifts in feature and label distribution, imbalanced user dataset sizes, drift in feature distribution conditioned on the labels and shift in the labeling function itself. 
This is caused by the independent and diverse user-specific contexts that predicate the data generation process.
For example, the style and content of a written message may differ depending on the author's age, culture, and geographical location. Indeed such heterogeneity can be seen in text datasets commonly used as benchmarks (see Section~\ref{sec:related-work}).

However the image domain suffers from a limited selection of large-scale datasets with realistic user partitions to benchmark algorithms and models (see Section \ref{sec:related-work}).
When new hypotheses are tested, researchers typically use centrally available data to simulate the federated setting.
For example, many works are evaluated by repurposing traditional benchmarks, such as MNIST \citep{lecun1998mnist} and CIFAR10 \citep{krizhevsky2009learning}, by creating artificial user partitions~\cite{hsu2019measuring}.
It is unclear if such artificial partitions are realistic enough to give confidence that hypotheses evaluated on these will transfer to federated learning in a real-world scenario.

\input{assets/explore_image}

We introduce FLAIR, a large-scale multi-label image classification dataset, for benchmarking federated learning algorithms and models. The datast has a total of 429,078 images originating from Flickr~\citep{flickr} and partitioned by 51,414 real user IDs. The images are annotated with labels from a two-level hierarchy, allowing us to define benchmarks with two levels of difficulty: the easier task has $17$ coarse-grained classes and the harder task has 1,628 fine-grained classes.
FLAIR also inherits many of the aforementioned non-IID characteristics:

\begin{itemize}
\item Imbalanced partitions --- Users have different number of images. The majority of users have only 1-10 images, but the most active users have hundreds of images. 
\item Feature distribution skew --- Users have different cameras, camera settings, which affect pixel generation.
\item Label distribution skew --- Users take photos of objects that align with their interests, which vary across photographers.
\item Conditional feature distribution skew --- Photos of the same category of objects can look very different due to weather conditions, cultural and geographical differences.
\item Label shifts --- There are no significant discrepancies in how the labels are used. However, the label generating process consisted of multiple human annotators, which may result in slight differences in how labeling decisions were made.
\end{itemize}

We provide benchmarks and analyze the performance of different settings of interest for FLAIR: centralized learning; federated learning; federated learning with central differential privacy; using random initialization of model parameters; and using model parameters pretrained on ImageNet~\citep{recht2019imagenet}.

%% file: assets/explore_image.tex
\begin{figure}[t]
\centering
\begin{tcolorbox}[title={\centering Flickr user 9334511@N06},arc=0mm,boxsep=0pt,left=-5pt,right=-5pt,top=1pt,bottom=1pt,width=\textwidth,halign=center,colback=white]
\begin{subfigure}[t]{.13\textwidth}
\includegraphics[width=\linewidth]{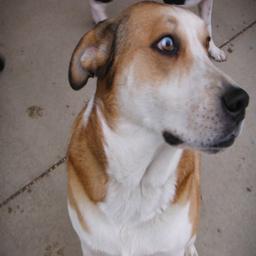}
\caption*{jack russell \\ terrier,\\land}
\end{subfigure}
\begin{subfigure}[t]{.13\textwidth}
\includegraphics[width=\linewidth]{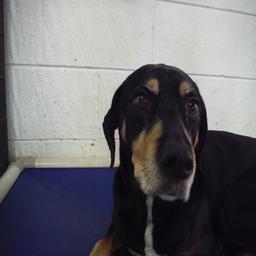}
\caption*{dog,\\material,\\structure}
\end{subfigure}
\begin{subfigure}[t]{.13\textwidth}
\includegraphics[width=\linewidth]{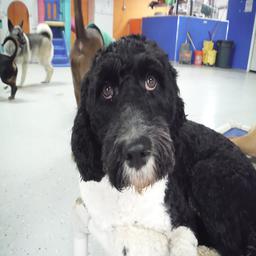}
\caption*{dog,\\interior room,\\structure}
\end{subfigure}
\begin{subfigure}[t]{.13\textwidth}
\includegraphics[width=\linewidth]{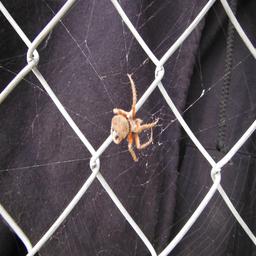}
\caption*{raw metal,\\spider,\\spiderweb}
\end{subfigure}
\begin{subfigure}[t]{.13\textwidth}
\includegraphics[width=\linewidth]{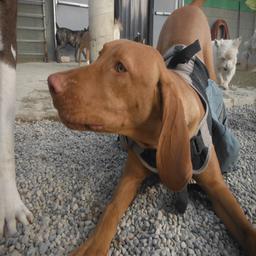}
\caption*{clothing,\\pillar,\\rocks,\\terrier}
\end{subfigure}
\begin{subfigure}[t]{.13\textwidth}
\includegraphics[width=\linewidth]{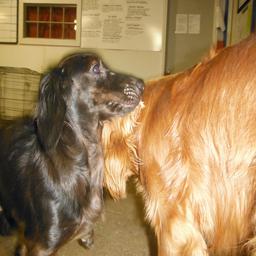}
\caption*{cage,\\dachshund,\\document,\\door}
\end{subfigure}
\begin{subfigure}[t]{.13\textwidth}
\includegraphics[width=\linewidth]{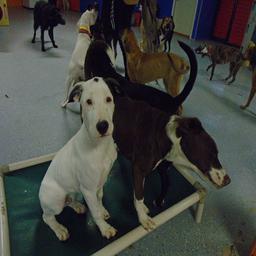}
\caption*{equipment,\\interior room,\\material,\\terrier}
\end{subfigure}
\end{tcolorbox}

\begin{tcolorbox}[title={\centering Flickr user 129851880@N07},arc=0mm,boxsep=0pt,left=-5pt,right=-5pt,top=1pt,bottom=1pt,width=\textwidth,halign=center,colback=white]
\begin{subfigure}[t]{.13\textwidth}
\includegraphics[width=\linewidth]{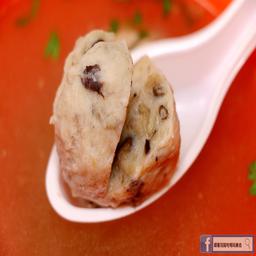}
\caption*{food,\\logo other,\\soup,\\spoon}
\end{subfigure}
\begin{subfigure}[t]{.13\textwidth}
\includegraphics[width=\linewidth]{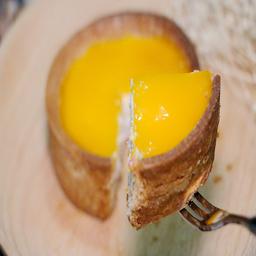}
\caption*{baked goods,\\fork}
\end{subfigure}
\begin{subfigure}[t]{.13\textwidth}
\includegraphics[width=\linewidth]{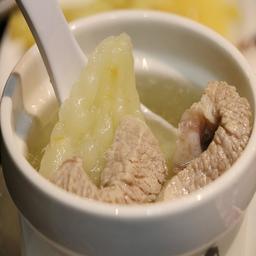}
\caption*{bowl,\\food,\\meat,\\soup}
\end{subfigure}
\begin{subfigure}[t]{.13\textwidth}
\includegraphics[width=\linewidth]{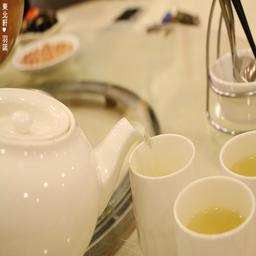}
\caption*{bowl,\\cup,\\ladle,\\material}
\end{subfigure}
\begin{subfigure}[t]{.13\textwidth}
\includegraphics[width=\linewidth]{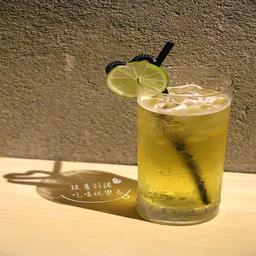}
\caption*{drink,\\glass,\\lime,\\material}
\end{subfigure}
\begin{subfigure}[t]{.13\textwidth}
\includegraphics[width=\linewidth]{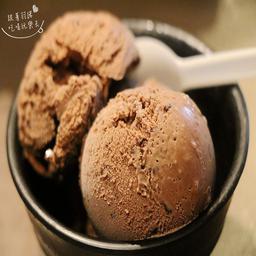}
\caption*{container,\\ice cream,\\spoon}
\end{subfigure}
\begin{subfigure}[t]{.13\textwidth}
\includegraphics[width=\linewidth]{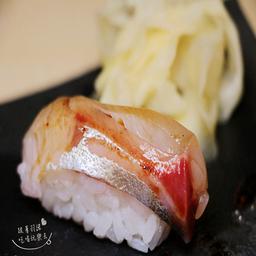}
\caption*{food,\\meat,\\rice}
\end{subfigure}
\end{tcolorbox}

\begin{tcolorbox}[title={\centering Flickr user 29694550@N06},arc=0mm,boxsep=0pt,left=-5pt,right=-5pt,top=1pt,bottom=1pt,width=\textwidth,halign=center,colback=white]
\begin{subfigure}[t]{.13\textwidth}
\includegraphics[width=\linewidth]{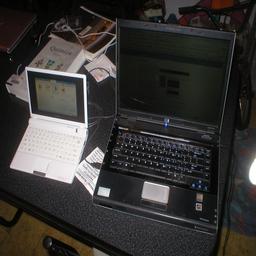}
\caption*{bicycle,\\book,\\laptop,\\printed page}
\end{subfigure}
\begin{subfigure}[t]{.13\textwidth}
\includegraphics[width=\linewidth]{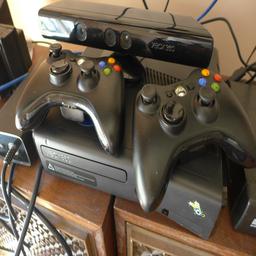}
\caption*{electronics,\\gamepad,\\joystick,\\wire}
\end{subfigure}
\begin{subfigure}[t]{.13\textwidth}
\includegraphics[width=\linewidth]{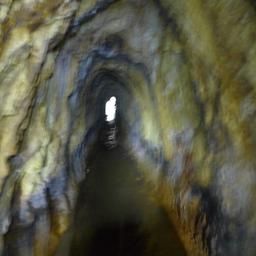}
\caption*{arch,\\cave,\\waterways}
\end{subfigure}
\begin{subfigure}[t]{.13\textwidth}
\includegraphics[width=\linewidth]{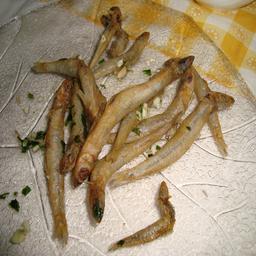}
\caption*{herb other,\\material,\\seafood,\\tableware}
\end{subfigure}
\begin{subfigure}[t]{.13\textwidth}
\includegraphics[width=\linewidth]{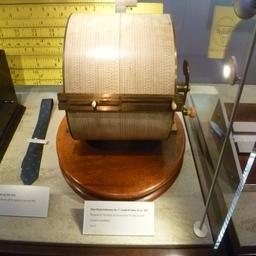}
\caption*{document,\\raw glass,\\structure,\\textile}
\end{subfigure}
\begin{subfigure}[t]{.13\textwidth}
\includegraphics[width=\linewidth]{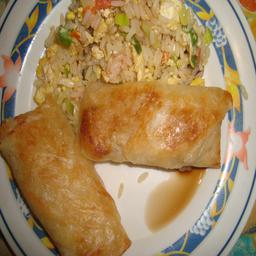}
\caption*{baked goods,\\plate,\\rice}
\end{subfigure}
\begin{subfigure}[t]{.13\textwidth}
\includegraphics[width=\linewidth]{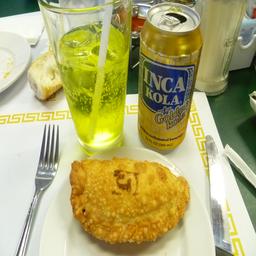}
\caption*{baked goods,\\calzone,\\container,\\drink}
\end{subfigure}
\end{tcolorbox}

\begin{tcolorbox}[title={\centering Flickr user 40164909@N00},arc=0mm,boxsep=0pt,left=-5pt,right=-5pt,top=1pt,bottom=1pt,width=\textwidth,halign=center,colback=white]
\begin{subfigure}[t]{.13\textwidth}
\includegraphics[width=\linewidth]{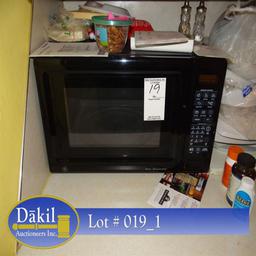}
\caption*{bag,\\container,\\document,\\material}
\end{subfigure}
\begin{subfigure}[t]{.13\textwidth}
\includegraphics[width=\linewidth]{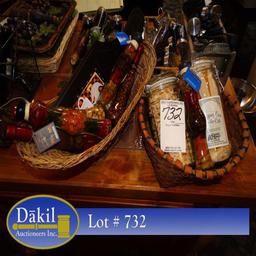}
\caption*{basket,\\bottle,\\document,\\jar}
\end{subfigure}
\begin{subfigure}[t]{.13\textwidth}
\includegraphics[width=\linewidth]{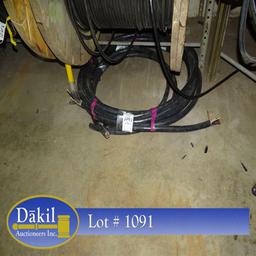}
\caption*{cord,\\equipment,\\logo,\\material}
\end{subfigure}
\begin{subfigure}[t]{.13\textwidth}
\includegraphics[width=\linewidth]{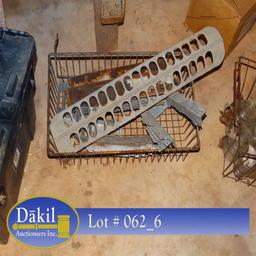}
\caption*{cardboard,\\container,\\logo,\\material}
\end{subfigure}
\begin{subfigure}[t]{.13\textwidth}
\includegraphics[width=\linewidth]{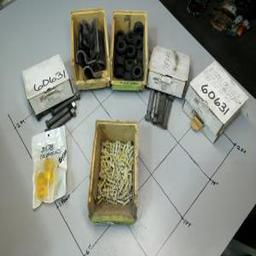}
\caption*{bag,\\carton,\\container,\\document}
\end{subfigure}
\begin{subfigure}[t]{.13\textwidth}
\includegraphics[width=\linewidth]{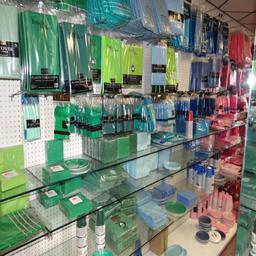}
\caption*{bag,\\container,\\interior shop,\\raw glass}
\end{subfigure}
\begin{subfigure}[t]{.13\textwidth}
\includegraphics[width=\linewidth]{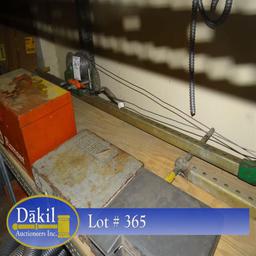}
\caption*{container,\\hose,\\logo,\\pulley}
\end{subfigure}
\end{tcolorbox}


\caption{FLAIR sample images and labels. Images in the same row are from the same Flickr user. Captions below each image are the annotated fine-grained labels.}
\label{fig:image_examples}
\end{figure}

%% file: related.tex
\section{Related Work}
\label{sec:related-work}
Previous work has mainly used two methods for preparing federated datasets: artificial partitioning of existing open-source datasets not originally purposed for federated learning~\cite{hsu2019measuring}, and 
constructing realistic partitions using real user identifiers preserved from the data generation process~\cite{mcmahan2018learning}.
The former approach requires fewer resources but more assumptions, and has been used with MNIST~\cite{lecun1998mnist}, CIFAR~\cite{krizhevsky2009learning}, and CelebA~\cite{liu2015faceattributes} datasets.
These experimental setups rely on artificially inducing some of the characteristics of real federated datasets during the sampling process.
Pachinko allocation based sampling method  was proposed to generate more realistic heterogeneous partitions, but it requires a hierarchy of coarse labels such as present in CIFAR100 \citep{reddi2020adaptive}.
Yet, there is no clear way for measuring how realistic the splits are.
In fact, federated data partitions in the wild are usually more heavy tailed than the artificial partitions previous work has used (see Section \ref{sec:dataset-statistics}).

The latter approach relies on datasets generated from a collection of users, with the user identifiers preserved.
Previous works have extensively used text datasets that have this property: Sentiment140~\cite{go2009twitter}, Shakespeare \citep{mcmahan2017communication}, Reddit\citep{caldas2018leaf} and StackOverflow \citep{stackoverflow}.
Realistic image datasets commonly used in the federated learning community are EMNIST \citep{caldas2018leaf}, iNaturalist-User-120k and Landmarks-User-160k \citep{hsu2020federated}. The landmarks dataset is the largest of them, with $164,172$ images, but has only $1262$ users, making it ill-suited for large-scale federated learning, especially for private federated learning where large batch sizes are typically needed.

Meta-learning is a ML paradigm closely related to federated learning, hence requiring similar kinds of datasets. 
Popular image datasets for meta-learning include Mini-Imagenet \citep{vinyals2016matching}, CUB-200-2011 \citep{wah2011caltech} and Omniglot \citep{lake2015human}.
These datasets are relatively small and low-resolution, with either artificial task partitioning or easy tasks, e.g. the original model-agnostic meta-learning algorithm already achieves $99.9\%$ accuracy with 5-way 5-shot classification on Omniglot \citep{finn2017model}.

Testing a hypothesis with a standardized benchmark agreed upon by the research community is essential for systematically making progress in the field of machine learning.
There are several benchmark suites that attempt to do this for federated learning: LEAF \citep{caldas2018leaf}, FedML \citep{he2020fedml}, OARF \citep{hu2020oarf} and FedScale \citep{lai2021fedscale}.
FedScale proposes a benchmark for image classification on Flickr images, which is similar to FLAIR.
This however is a multiclass dataset, where image-label pairs are constructed by cropping single objects from bounding box annotations; this results in many duplicate images with different labels because the bounding boxes commonly overlap.
As explored more thoroughly in Section \ref{sec:label-statistics}, FLAIR also has a more diverse set of classes and includes two levels of difficulty.

%% file: prelim.tex
\section{Preliminaries}


{\bf Federated learning}~\cite{mcmahan2017communication} enables training on users' data without collecting or storing the data on a centralized server.  
In each round of federated learning, the server samples a cohort of users and sends the current model to the sampled users' devices. 
The sampled users train the model locally with SGD and share the gradient updates back to the server after local training. 
The server updates the global model, treating the aggregate of the per-user updates in lieu of a gradient estimate in an optimization algorithm such as  SGD or Adam~\cite{kingma2015adam,reddi2020adaptive}. 

\noindent{\bf Differential Privacy.} 
Even though user data is not shared with the server in the federated setting, the shared gradient updates can still reveal sensitive information about user data~\cite{melis2019exploiting,zhu2019deep}. Differential privacy (DP)~\cite{dwork2006calibrating} can be used to provide a formal privacy guarantee to prevent such data leakage in the federated setting. 

\begin{definition}[Differential privacy]
A randomized mechanism $\gM: \gD \mapsto \gR$ with a domain $\gD$ and range $\gR$ satisfies $(\epsilon, \delta)$-differential privacy if for any two adjacent datasets $d, d^\prime \in \gD$ and for any subset of outputs $S \subseteq \gR$ it holds that $\prob[M(d) \in S] \leq e^\epsilon \prob[\gM(d^\prime) \in S] + \delta$.
\end{definition}

In the context of DP federated learning~\cite{mcmahan2018learning}, $\gD$ is the set of all possible datasets with examples associated with users, range $\gR$ is the set of all possible models, and two datasets $d, d^\prime$ are adjacent if  $d^\prime$ can be formed by adding or removing all of the examples associated with a single user from $d$.

When a federated learning model is trained with DP, the model distribution is close to what it would be if a particular user did not participate in the training. 
Following prior works in DP-SGD~\cite{abadi2016deep} in the federated learning context~\cite{mcmahan2018learning}, 
two modifications are made to the federated learning algorithm to provide a DP guarantee: 1) model updates from each user are clipped so that their $L_2$ norm is bounded, and 2) Gaussian noise is added to the aggregated model updates from all sampled users. 
For the purpose of privacy accounting, we assume that each cohort is formed by sampling each user uniformly and independently, and that this sample is hidden from the adversary.

%% file: data.tex
\section{FLAIR Dataset}


\subsection{Dataset collection}
The initial set of images was curated with the Flickr API~\footnote{\url{https://www.flickr.com/services/api/}}.
The corresponding Flickr user IDs were preserved so that the images were naturally grouped by users.
All curated images are publicly shared by the Flickr users and permissively licensed (detailed in Appendix~\ref{app:license}). 

\noindent{\bf Filtering.}
We enforce strict filtering criteria to remove images that may contain personally identifiable information (PII).
We use a two stage filtering approach: 1) we apply a face detection model to automatically remove images with faces, and; 2) we rely on human annotators to filter the remaining images that contains PII. 
Specifically, two annotators were assigned for filtering each image where the first annotator flags whether an image contains PII and the second annotator validates the results.
See Appendix~\ref{app:filtering} for detailed filtering guideline.

\noindent{\bf Annotation.}
\label{sec:annotation}
The images from Flickr API were initially unlabeled. 
We annotated the images with the main objects present in the images using a taxonomy of 1,628 fine-grained classes. 
We also defined 17 coarse-grained classes in the taxonomy, where each fine-grained class is associated with a coarse-grained class.
Similar to filtering, two annotators were assigned for labeling and validating each image. 
If there was an ambiguous object present in the image and the annotator could not tell which fine-grained label to assign, a coarse-grained label was added instead.

\subsection{FLAIR dataset statistics}
\label{sec:dataset-statistics}
After filtering and annotation, the finalized FLAIR dataset contain 429,078 images from 51,414 Flickr users, with 17 coarse-grained labels and 1,628 fine-grained labels.

\noindent{\bf User data statistics.}
\begin{figure}[t]
\centering
\begin{subfigure}[t]{.49\textwidth}
\includegraphics[width=\linewidth]{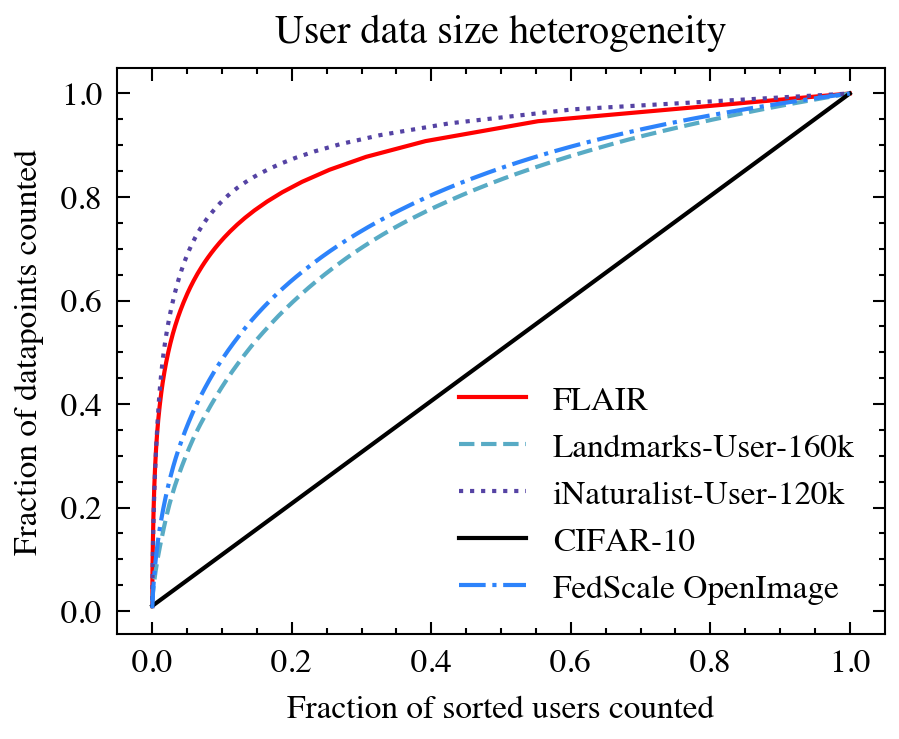}
\end{subfigure}
\begin{subfigure}[t]{.49\textwidth}
\includegraphics[width=\linewidth]{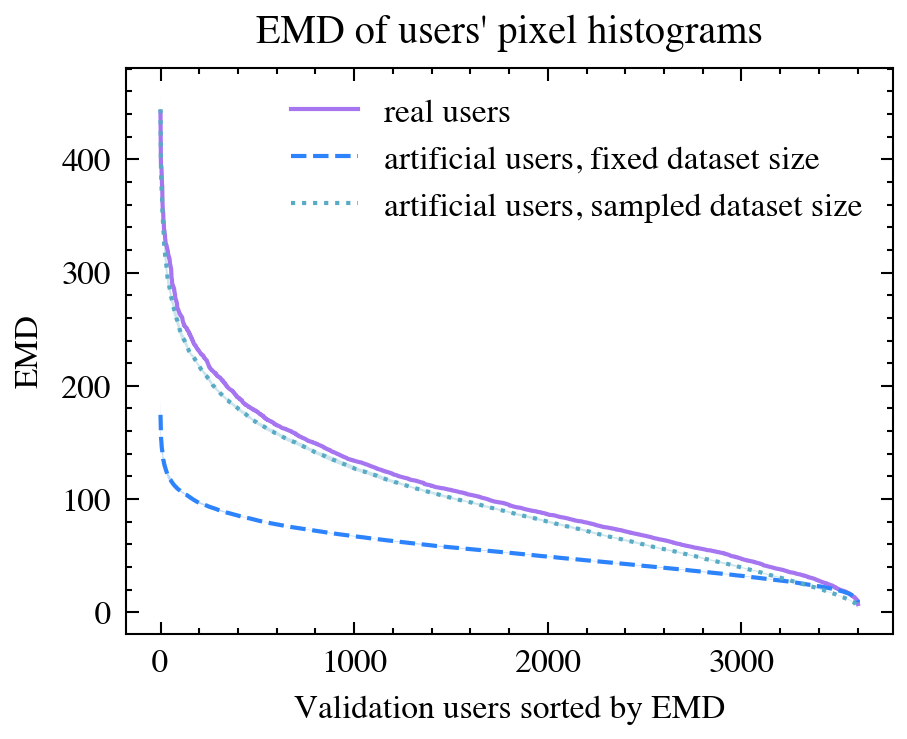}
\end{subfigure}

\caption{\textbf{Left:} Cumulative dataset length of users in descending order of quantity, normalized by number of users on x-axis and number of datapoints on y-axis.
\textbf{Right:} Earth mover's distance (EMD) between users pixel histogram and the overall average pixel histogram from class \emph{structure}. Blue line is from simulating user splits with equal dataset size, green line is simulating user splits by sampling from the real dataset size distribution, and purple line is the actual split.}
\label{fig:image_pixel_similarity}
\end{figure}
The number of images per user is significantly skewed, where the largest $2.3\%$ of users collectively have as many images as the bottom $97.7\%$ of users. The left of Figure \ref{fig:image_pixel_similarity} compares the quantity skew for FLAIR and other image classification benchmarks for federated learning.
In the case of CIFAR, there is a straight line because there is no skew.
The figure indicates that FLAIR has the second largest quantity skew, after iNaturalist-User-120k.

To visualize the feature distribution skew in FLAIR, we show in Figure~\ref{fig:image_pixel_similarity} (right) the earthmover distance (EMD) between the average pixel histogram of a user's images, to the population average pixel histogram. 
EMD is computed on the images from the most common label, \emph{structure}, to remove any skew that the class imbalance might cause.
The quantity skew also causes feature distribution skew (comparing blue line to green line), and the real non-iid partitioning slightly increases the skew compared to the average simulated non-iid partitioning (comparing green line to purple line).

\noindent{\bf Label statistics.}
\label{sec:label-statistics}
%
%
\begin{figure}[t]
\centering
\begin{subfigure}[t]{.49\textwidth}
\includegraphics[width=\linewidth]{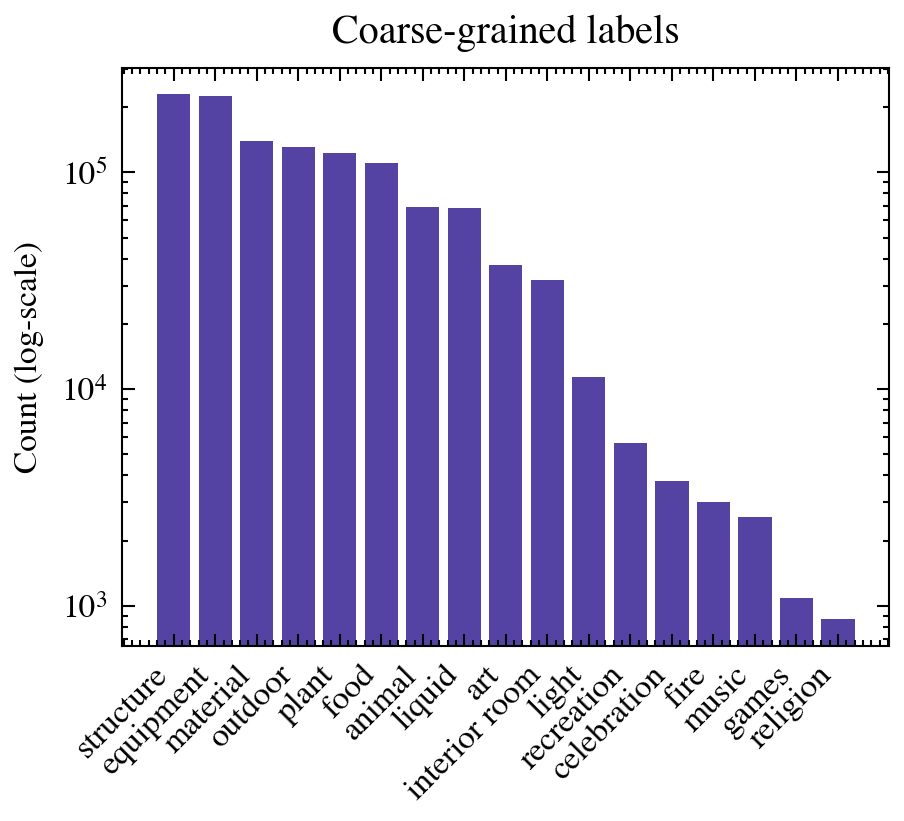}
\end{subfigure}
\begin{subfigure}[t]{.49\textwidth}
\savebox{\tempfig}{\includegraphics[width=\linewidth]{assets/dataset-stats/coarse-label-dist.png}}
\raisebox{\dimexpr\ht\tempfig-\height}{\includegraphics[width=\linewidth]{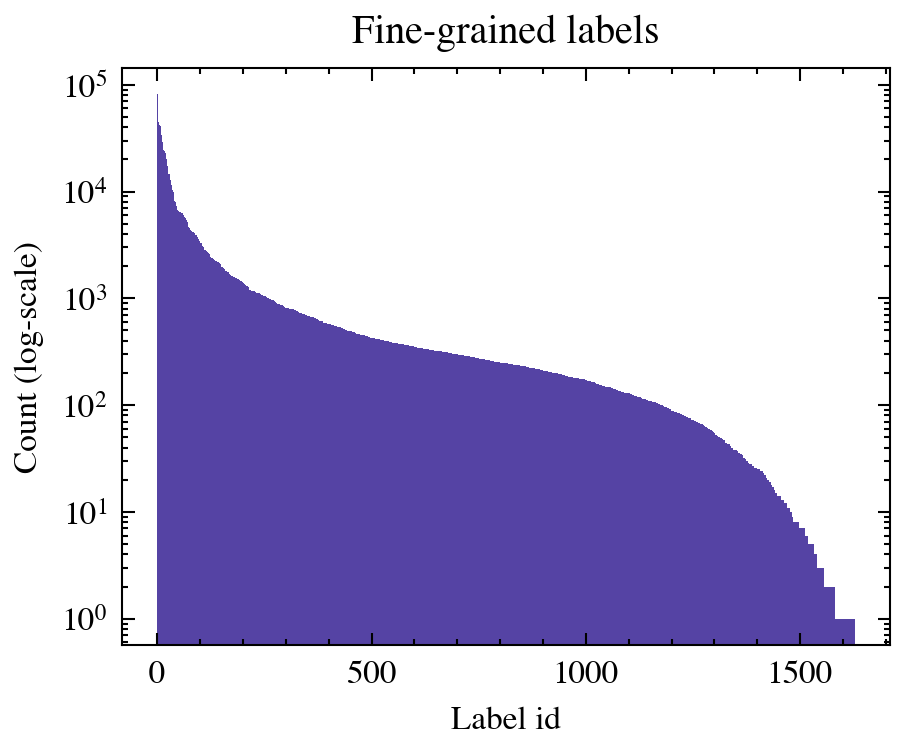}}
\end{subfigure}

\caption{FLAIR label distribution for coarse-grained and fine-grained taxonomies. }
\label{fig:label_dist}
\end{figure}
Figure \ref{fig:label_dist} shows the total count of all labels across all users, revealing a significant class imbalance.
The most common coarse-grained class, \emph{structure}, occurs 228,923 times on a total of 87\% of users. The least common coarse-grained class, \emph{religion}, occurs 866 times on a total of 1.4\% of users.
The fine-grained labels similarly have a skewed distribution, with 1255 out of the 1628 classes being present on less than $0.1\%$ of users.

\noindent{\bf Dataset split.}
For comparable and reproducible benchmarks on the FLAIR dataset, we provide a train-test split based on Flickr user IDs, such that the data of a particular user is only present in one of three partitions of the data.
Out of 51,414 Flickr users, $80\%$ are in the training set, $10\%$ in the validation set and $10\%$ in the test set. 
There are 345,879 images in total in the training set, 39,239 in the validation set and 43,960 in the test set.

%% file: experiments.tex
\section{Experiments}
\subsection{Benchmark setups}
\label{sec:setup}

\noindent{\bf Learning settings.} We benchmark the FLAIR dataset in three different learning settings: centralized learning, non-private and private federated learning. Comparing these settings demonstrates how heterogeneity of the user data distribution and providing user privacy guarantees affect model convergence. In the centralized learning setting, training data is the union of images from all users in the training split and user ID is ignored. 

\noindent{\bf ML tasks and models.}
As described in Section~\ref{sec:annotation}, the main objects in each image were annotated into coarse-grained and fine-grained taxonomies.
We consider multi-label classification task on these two taxonomies, i.e. predicting if a class is presented in an image for each class in the taxonomy. 

We use a ResNet-18~\cite{he2016deep} model for all benchmark experiments.
The final classification layer is a 17-way logistic regression for the coarse-grained taxonomy and 1,628-way for the fine-grained.
The model has more than 11M parameters in total.
We consider both training from scratch (i.e. from a random initialization) and fine-tuning from a pretrained model.
The pretrained ResNet-18 model is obtained from the Torchvision repository~\footnote{\url{https://github.com/pytorch/vision}} and was trained on the ImageNet dataset~\cite{deng2009imagenet}.

For models trained from scratch, we further replace all batch normalization (BN)~\cite{ioffe2015batch} layers with group normalization~\cite{wu2018group} to avoid sharing the sensitive states in BN with the server in federated settings. 
For the pretrained ResNet-18 model, we freeze the BN states during fine-tuning and only update the scale and bias parameters.

\noindent{\bf Evaluation metrics.}
We use standard multi-label classification metrics for the benchmark, including precision (percentage of predicted objects that are actually in the images), recall (percentage of objects in the images are predicted), F1 score, and averaged precision (AP) score. We report overall (micro-averaged) metrics, obtained by averaging over all examples, and per-class (macro-averaged) metrics, obtained by taking the average over classes of the average over examples restricted to a specific class. 

\noindent{\bf Simulating large cohort noise-level with small cohort.}
When training with DP, increasing cohort size $C$ will monotonically increase the signal-to-noise ratio (SNR) of the averaged noisy aggregates as the DP noise will be reduced by averaging. 
As we will show later in Section~\ref{sec:results}, the minimum SNR required for training large neural networks such as ResNet corresponds to a $C$ in the thousands, which is  compute-intensive with current federated learning frameworks.

Following prior work~\cite{mcmahan2018learning}, we simulate the SNR effect of a large cohort $C_\mathrm{lg}$ using a small cohort $C_\mathrm{sm}$ so that we can efficiently experiment with different noise-levels.
Let $\sigma = \gM(\cdot, C)$ be the noise multiplier calculated by moments accountant $\gM$~\cite{abadi2016deep} for cohort size $C$ and other privacy parameters. 
We use $C_\mathrm{sm}$ and noise multiplier $\sigma_\mathrm{sm}$ for experiments, where
$\sigma_\mathrm{sm} = \frac{C_\mathrm{sm}}{C_\mathrm{lg}}\gM(\cdot, C_\mathrm{lg}).$
The noise applied to the averaged $C_\mathrm{sm}$ model updates has standard deviation $\frac{\sigma_\mathrm{sm}}{ C_\mathrm{sm}} = \frac{\gM(\cdot, C_\mathrm{lg})}{ C_\mathrm{lg}}$, which is the same as if we are training with $C_\mathrm{lg}$ users.

\noindent{\bf Hyperparameters.}
For all experiments, we use Adam~\cite{kingma2015adam,reddi2020adaptive}
as the server-side optimizer.
During training, each image is randomly cropped to size $224\times224$ and randomly flipped horizontally or vertically.
During evaluation, each image is resized to $224\times224$.
We performed a grid search on the hyperparameters and report the values that yield best performance on the validation set. See Appendix~\ref{app:hyper} for hyperparameters grids.

For the centralized setting, we set the number of epochs to be 100 and the learning rate to be 5e-4 if training from scratch, and number of epochs to be 50 and learning rate to be 1e-4 when fine-tuning. We use a mini-batch size of 512.

For the federated learning setting, we train the model for 5,000 rounds with a cohort size of 200.
We set the server learning rate  to 0.1. 
Each sampled user trains the model locally with SGD for 2 epochs with local batch size set to 16 and local learning rate set to 0.1 when training from scratch and 0.01 when fine-tuning.
We limit the maximum number of images for each user to be 512 and if a user has more images, we randomly sample 512 images to train.

For federated learning with differential privacy, we use $\epsilon=2.0, \delta=N^{-1.1}$ where $N$ is the number of training users.
We set the server learning rate  to 0.02. 
We use an adaptive clipping algorithm~\cite{andrew2021differentially} to tune the clipping bound, with the $L_2$ norm quantile set to 0.1.
We use 200 users sampled per round to simulate the noise-level with a cohort size of 5,000, and we also analyze the effect of different cohort sizes in Section~\ref{sec:results}.

\subsection{Results}
\label{sec:results}

\begin{table}[t]
\centering
\footnotesize
\caption{FLAIR benchmark results on test set. For setting, C, FL, PFL stands for centralized, federated and private federated learning. C and O denotes whether the metrics are per-class or overall. AP denotes averaged precision; P denotes precision; R denotes recall; and F1 denotes F1 score.}
\begin{tabular}{lll|rrrr|rrrr}
\toprule
Setting & Init & Label & C-AP & C-P & C-R & C-F1 & O-AP & O-P & O-R & O-F1\\
\midrule
C & Random & Coarse & 60.40 & 72.79 & 48.24 & 58.03 & 87.61 & 81.43 & 75.06 & 78.11 \\
FL & Random & Coarse & 50.41 & 59.74 & 37.46 & 46.04 & 82.87 & 78.25 & 69.02 & 73.35 \\
PFL & Random & Coarse & 28.80 & 30.02 & 17.85 & 22.39 & 63.19 & 68.67 & 43.42 & 53.20 \\
\midrule
C & ImageNet & Coarse & 67.71 & 75.71 & 55.42 & 64.00 & 90.40 & 84.09 & 78.96 & 81.44 \\
FL & ImageNet & Coarse & 62.09 & 71.81 & 48.60 & 57.97 & 88.77 & 83.50 & 75.95 & 79.54 \\
PFL & ImageNet & Coarse & 44.28 & 47.25 & 32.30 & 38.37 & 80.20 & 77.51 & 64.37 & 70.33 \\
\midrule
C & Random & Fine & 14.90 & 26.25 & 7.18 & 11.27 & 43.15 & 66.38 & 26.02 & 37.38 \\
FL & Random & Fine & 1.53 & 0.91 & 0.28 & 0.43 & 22.68 & 58.99 & 8.38 & 14.68 \\
PFL & Random & Fine & 0.29 & 0.00 & 0.00 & 0.00 & 7.03 & 0.00 & 0.00 & 0.00 \\
\midrule
C & ImageNet & Fine & 20.26 & 32.97 & 10.92 & 16.40 & 47.95 & 68.73 & 30.04 & 41.81 \\
FL & ImageNet & Fine & 2.03 & 1.99 & 0.40 & 0.66 & 27.31 & 65.47 & 10.50 & 18.10 \\
PFL & ImageNet & Fine & 0.53 & 0.22 & 0.01 & 0.01 & 12.67 & 57.01 & 0.27 & 0.54 \\
\bottomrule
\end{tabular}
\label{tab:results}
\end{table}

Table~\ref{tab:results} summarizes the benchmark results on the FLAIR test set. 
For the coarse-grained taxonomy, we observe that the performance gap between centralized and federated setting is about 20\% on the per class metrics and 6\% on the overall metrics if the models are trained from scratch.
These gaps are reduced to 8\% and 2\% if models are fine-tuned from pretrained ResNet.
When DP is applied, the per class metrics drop about 40\% and overall metrics 24\% from non-private federated learning if training from scratch.
When fine-tuning with DP, the drop is less significant, about 30\% for per class metrics and 10\% for overall metrics.

For the fine-grained taxonomy, federated learning performance is much worse than the centralized baseline. 
The gaps are around 90\% and 50\% for per-class and overall metrics regardless whether the model is trained from scratch or started from a pretrained model. 
DP model has even worse performance compared to non-private one due to the extra noise introduced, which indicates long-tailed prediction tasks are especially hard in private federated learning setting due to the sparse label distribution among users.
 
\begin{table}[t]
\caption{Averaged precision for each coarse-grained class. C, FL, PFL stands for centralized, federated and private federated learning. R and F stands for training from scratch and fine-tuning. Columns are sorted by decreasing order of class frequency.}

\setlength{\tabcolsep}{3pt}
\newcommand{\twidth}{0.04\linewidth}

\centering
\scriptsize
\begin{tabular}{p{0.05\linewidth}|p{\twidth}p{\twidth}p{\twidth}p{\twidth}p{\twidth}p{\twidth}p{\twidth}p{\twidth}p{\twidth}p{\twidth}p{\twidth}p{\twidth}p{\twidth}p{\twidth}p{\twidth}p{\twidth}p{\twidth}}
\toprule
Setting & struc-ture & equip-ment & mate-rial & out-door & plant & food & animal & liquid & art & interior room & light & recrea-tion & celeb-ration & fire & music & games & reli-gion 
\\
\midrule
C-R & 90.1 & 92.8 & 66.9 & 95.1 & 93.0 & 95.0 & 87.0 & 78.6 & 43.0 & 65.5 & 35.6 & 30.2 & 36.2 & 63.4 & 14.8 & 19.6 & 19.9\\
FL-R & 86.0 & 90.0 & 61.3 & 92.6 & 89.7 & 91.1 & 75.0 & 67.5 & 31.0 & 56.7 & 25.8 & 17.4 & 19.4 & 37.8 & 5.3 & 2.3 & 8.2\\
PFL-R & 65.9 & 73.5 & 43.1 & 77.3 & 65.9 & 67.7 & 26.1 & 34.6 & 10.8 & 12.8 & 5.3 & 2.5 & 1.2 & 1.7 & 0.7 & 0.3 & 0.2\\
\midrule
C-F & 92.5 & 94.6 & 70.8 & 96.3 & 94.0 & 96.6 & 93.5 & 84.5 & 55.5 & 71.4 & 40.8 & 41.1 & 46.5 & 70.3 & 39.3 & 42.4 & 21.1\\
FL-F & 91.2 & 93.6 & 68.8 & 95.6 & 92.7 & 95.7 & 90.2 & 79.5 & 48.6 & 68.4 & 34.5 & 18.3 & 35.0 & 61.1 & 15.5 & 17.0 & 7.6\\
PFL-F & 84.1 & 88.4 & 56.3 & 89.6 & 86.3 & 89.8 & 76.8 & 56.7 & 25.1 & 52.5 & 17.9 & 4.0 & 3.9 & 18.6 & 2.0 & 0.2 & 0.7\\
\bottomrule
\end{tabular}
\label{tab:per_class_ap}
\end{table}

Table~\ref{tab:per_class_ap} summarizes averaged precision scores on FLAIR test set for each class in the coarse-grained taxonomy. 
The performances are different for different classes and there is a positive correlation between the frequency of the class and its performance. 
Noticeably, the gaps between classes are enlarged if models are trained with federated learning and DP. 
For instance, the gap between \textit{recreation} and \textit{outdoor} is about 68\% in centralized setting while the gap increases to 81\% in the federated setting and 96\% in the federated setting with DP.
In other words, the decrease in performance is worse for classes that are less frequent in federated learning, especially when DP is applied. This observation was also noted in prior works~\cite{bagdasaryan2019differential,song2019auditing}.

\begin{figure}[t]
\centering
\includegraphics[width=0.48\textwidth]{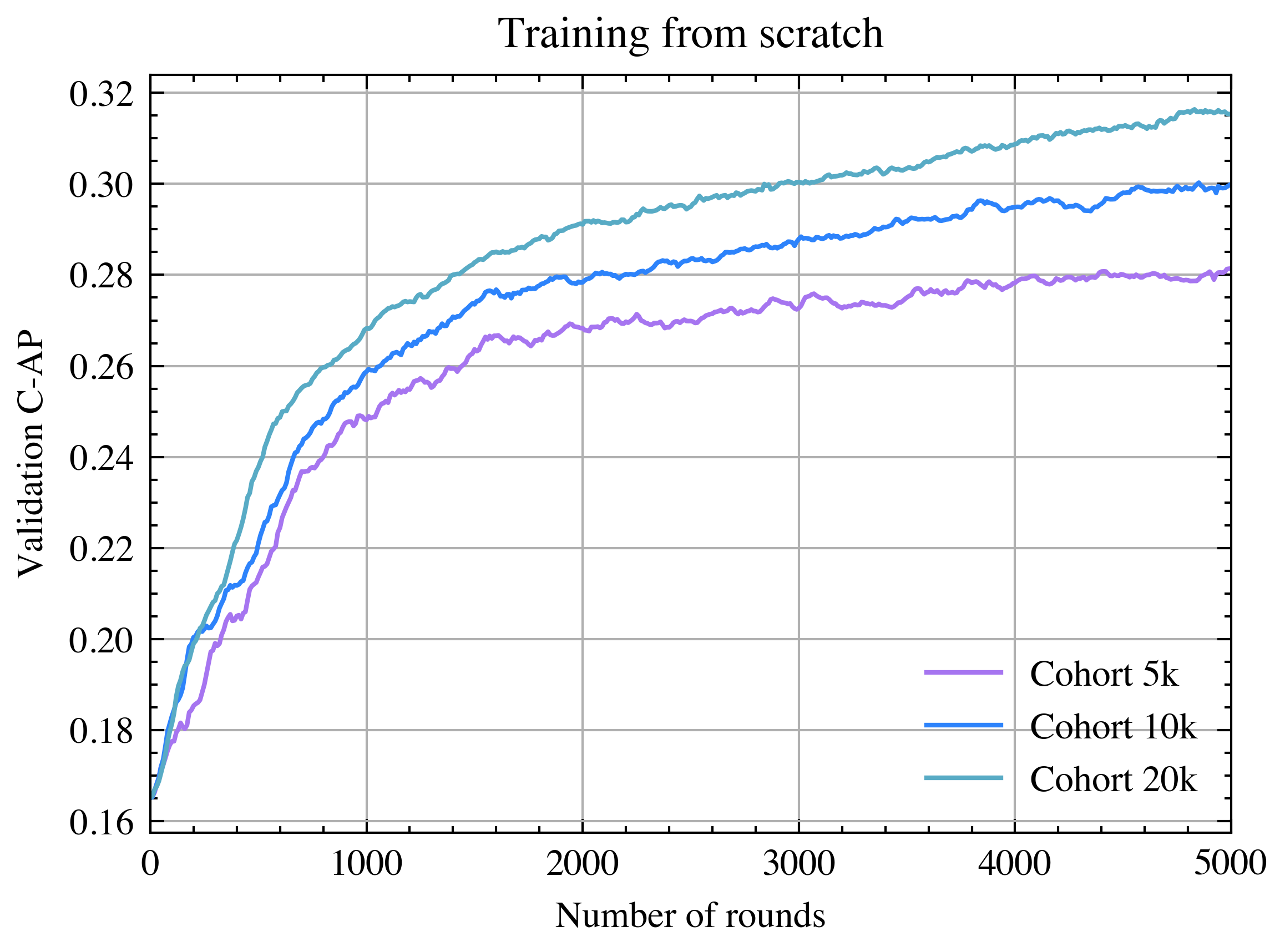}
\includegraphics[width=0.48\textwidth]{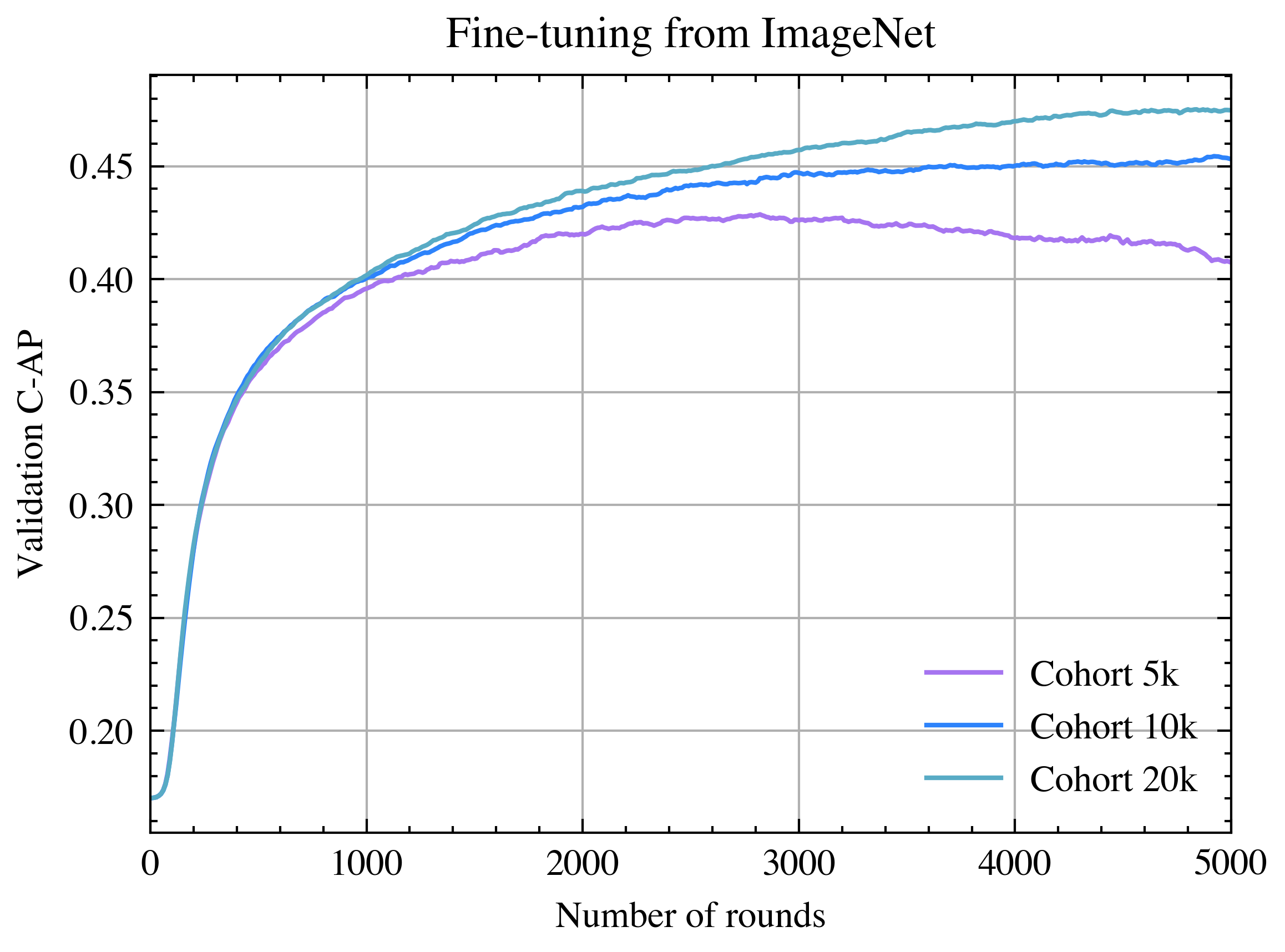}
\caption{Effect of cohort size in PFL training. x-axis is the number of rounds of federated learning and y-axis is the per-class AP on the validation set.}
\label{fig:cohort}
\end{figure}

\noindent{\bf Effect of cohort size on PFL.} 
As described in Section~\ref{sec:setup}, cohort size controls the noise-level of PFL, and thus we further examine the impact of cohort size on the performance of DP models.
Figure~\ref{fig:cohort} illustrates the per-class AP on the validation set in different rounds of PFL training.
For both training from scratch and fine-tuning, increasing cohort size yields faster and better generalization.

%% file: discussion.tex
\section{Discussion}
\subsection{Research directions}
\noindent{\bf Imbalanced classes.} For the coarse-grained taxonomy, models performed differently on different classes and the performance is correlated to the frequency of the class.
This difference is enlarged in federated learning, especially when DP is applied, indicating that the heterogeneity and DP noise worsened the imbalance problem. 
Given its heterogeneous nature, we believe FLAIR is a suitable dataset with which researchers can study the class imbalance problem in the distributed setting.

\noindent{\bf Few-shot and zero-shot federated learning.} As demonstrated in Section~\ref{sec:results}, federated learning models perform worse on FLAIR fine-grained taxonomy compared to coarse-grained.
Out of 1,628 fine-grained classes, 11 present in the validation and test dataset are unseen and 134 have less than 20 positive examples in the training set.
Predicting these few-shot and zero-shot labels can be very difficult even in the centralized training setting.
Indeed the signals for the tail classes in fine-grained taxonomy are extremely sparse and the sparsity is strengthened in federated learning as the infrequent classes are concentrated in only a few users.
Furthermore, DP exacerbates the performance of infrequent classes due to poor SNR of sparse gradients.
We believe the long-tailed label distribution in FLAIR fosters research interests in few-shot and zero-shot learning in the private federated setting.

\noindent{\bf Noise-robust and efficient federated learning with DP.} 
As shown in Figure~\ref{fig:cohort}, the larger the cohort size, the smaller the noise on the aggregated model updates and thus the better the model when trained with DP, especially for deep neural networks with tens of millions of parameters.
Larger cohorts increase the latency of federated learning with DP and may become impractical when the number of iterations required to converge is also large.
We believe the scale and complexity of FLAIR will inspire research in designing model architectures and optimization algorithms which are more robust to DP noise and also more efficient to train.

\noindent{\bf Personalization.} 
Personalization in federated learning is an active research area as a single model is unlikely to generalize equally well among all users.
Meta learning~\cite{fallah2020personalized} and local adaptation~\cite{deng2020adaptive,yu2020salvaging} are some of the attractive approaches for personalized federated learning.  
We did not benchmark FLAIR with personalization in this work and leave it for future works. 

\noindent{\bf Advanced vision models.}
As an initial benchmark, we only explored one model architecture, ResNet18. 
There are many more advanced architectures or pretrained models such as vision transformers~\cite{dosovitskiy2020image}, SimCLR~\cite{chen2020simple}, or CLIP~\cite{radford2021learning} that we did not use for experiments.
It is also an interesting research topic to search for the optimal model architectures in federated learning with DP. 

\subsection{Limitations}
Due to our strict filtering criteria, images with faces or identifiable human bodies are removed from FLAIR.
Thus, FLAIR is not suitable for any facial recognition or person identification vision tasks.
This filtering also reduced the size of the dataset, and may have changed the distributions of the number of images per user.

More generally, Federated Learning applications are diverse and various heterogeneity properties can vary a lot across applications. Any single dataset thus will not accurately represent all relevant properties of a specific application. Evaluating algorithms on a collection of datasets is thus important.

%% file: conclusion.tex
\section{Conclusions}
In this work, we presented FLAIR, a large-scale image dataset suitable for federated learning. 
We compared FLAIR with existing federated learning image datasets and discussed the advantages of FLAIR.
We described how the images in FLAIR were curated and annotated. 
We provided reproducible benchmarks for centralized, federated and differentially private settings.
We have open-sourced both the FLAIR dataset and the benchmark code for the community to use with the aim of in advancing the research in federated learning.

\subsection*{Acknowledgement}
We thank Flickr and the Flickr community for providing the set of images that made this dataset possible; Ulfar Erlingsson and Matt Seigel for their guidance during the early stages of this project; Yasmin Alameddine and Sophie Ostlund for their invaluable help throughout the project; multiple annotators for helping filter the dataset; Hanlin Goh and Aine Cahill for valuable feedback on the paper draft; Arjun Rangarajan, Plamena Gerovska, Katy Linksy, Wonhee Park, Vojta Jina, Mona Chitnis, Yulia Shuvkashvili, Julien Freudiger, Rogier van Dalen, Abhishek Bhowmick, Hillary Strickland, Subhash Sudan, Mya Exum, Laura Snarr, Guillaume Tartavel, Piotr Maj, Laurent Duchesne and Mark Faridani for their help with this effort.

%% file: appendix.tex
\newpage
\section{Datasheets for FLAIR Dataset}
\label{app:license}
\label{app:filtering}

\subsection{Motivation}

The questions in this section are primarily intended to encourage
dataset creators to clearly articulate their reasons for creating the
dataset and to promote transparency about funding interests.
The latter may be particularly relevant for datasets created for
research purposes.

\begin{itemize}

\item \textbf{For what purpose was the dataset created?} Was there a specific task in mind? Was there a specific gap that needed to be filled? Please provide a description.
\\
\answer{FLAIR dataset was created for the purpose of providing the community a benchmark in the vision domain to accelerate federated learning research.
FLAIR is suitable for multi-label image classification tasks, where the input is an image and output is a set of objects presented in the image.}

\item \textbf{Who created the dataset (e.g., which team, research group) and on behalf of which entity (e.g., company, institution, organization)?}
\\
\answer{Apple ML privacy team and ML research team created dataset on behalf of Apple Inc.}

\item \textbf{Who funded the creation of the dataset?} If there is an associated grant, please provide the name of the grantor and the grant name and number.
\\
\answer{Apple Inc.}


\end{itemize}

\subsection{Composition}

Dataset creators should read through {these questions} prior to
any data collection and then provide answers once {data} collection is
complete. Most of the questions {in this section} are intended to
provide dataset consumers with the information they need to make
informed decisions about using the dataset for their chosen
tasks. Some of the questions are {designed to elicit} information
about compliance with the EU's General Data Protection Regulation
(GDPR) or comparable regulations in other jurisdictions.

{Questions that apply only to datasets that relate to people are
grouped together at the end of the section. We recommend taking a
broad interpretation of whether a dataset relates to people. For
example, any dataset containing text that was written by people
relates to people.}

\begin{itemize}

\item \textbf{What do the instances that comprise the dataset
    represent (e.g., documents, photos, people, countries)?} Are there
  multiple types of instances (e.g., movies, users, and ratings;
  people and interactions between them; nodes and edges)? Please
  provide a description.
\\
\answer{The instances are Flickr images with annotation and metadata. }

\item \textbf{How many instances are there in total (of each type, if appropriate)?}
\\
\answer{There are 429,078 images in total.}

\item \textbf{Does the dataset contain all possible instances or is it
    a sample (not necessarily random) of instances from a larger set?}
  If the dataset is a sample, then what is the larger set? Is the
  sample representative of the larger set (e.g., geographic coverage)?
  If so, please describe how this representativeness was
  validated/verified. If it is not representative of the larger set,
  please describe why not (e.g., to cover a more diverse range of
  instances, because instances were withheld or unavailable).
\\
\answer{The instances in FLAIR dataset is a subset of the larger set, which is all Flickr images.
Not all Flickr images are suitable for research use, i.e. images with personal identifiable information and images without permissive license were excluded. }

\item \textbf{What data does each instance consist of?} ``Raw'' data
  (e.g., unprocessed text or images) or features? In either case,
  please provide a description.
 \\
\answer{Each instance consists of an image.}

\item \textbf{Is there a label or target associated with each
    instance?} If so, please provide a description.
\\ 
\answer{Each image has two sets of annotated labels from two taxonomies. Each image also has the associated Flickr user ID and image ID.}

\item \textbf{Is any information missing from individual instances?}
  If so, please provide a description, explaining why this information
  is missing (e.g., because it was unavailable). This does not include
  intentionally removed information, but might include, e.g., redacted
  text.
\\
\answer{No.}

\item \textbf{Are relationships between individual instances made
    explicit (e.g., users' movie ratings, social network links)?} If
  so, please describe how these relationships are made explicit.
\\
\answer{Yes, individual images from the same Flickr user have the same Flickr user ID.}

\item \textbf{Are there recommended data splits (e.g., training,
    development/validation, testing)?} If so, please provide a
  description of these splits, explaining the rationale behind them.
\\
\answer{Yes. 
FLAIR data is partitioned based on Flickr user IDs, such that the data of a particular user is present in only one of three splits.
Out of 51,414 Flickr users, $80\%$ are in the training set, $10\%$ in the validation set and $10\%$ in the test set. 
There are 345,879 images in total in the training set, 39,239 in the validation set and 43,960 in the test set.}

\item \textbf{Are there any errors, sources of noise, or redundancies
    in the dataset?} If so, please provide a description.
\\
\answer{N/A.} 

\item \textbf{Is the dataset self-contained, or does it link to or
    otherwise rely on external resources (e.g., websites, tweets,
    other datasets)?} If it links to or relies on external resources,
    a) are there guarantees that they will exist, and remain constant,
    over time; b) are there official archival versions of the complete
    dataset (i.e., including the external resources as they existed at
    the time the dataset was created); c) are there any restrictions
    (e.g., licenses, fees) associated with any of the external
    resources that might apply to a {dataset consumer}? Please provide
    descriptions of all external resources and any restrictions
    associated with them, as well as links or other access points, as
    appropriate.
\\
\answer{FLAIR is self-contained.}

\item \textbf{Does the dataset contain data that might be considered
    confidential (e.g., data that is protected by legal privilege or
    by doctor{--}patient confidentiality, data that includes the content
    of individuals' non-public communications)?} If so, please provide
    a description.
\\
\answer{No.}

\item \textbf{Does the dataset contain data that, if viewed directly,
    might be offensive, insulting, threatening, or might otherwise
    cause anxiety?} If so, please describe why.
\\
\answer{No. Images with offensive and other inappropriate materials have been removed from FLAIR.}

\end{itemize}

{If the dataset does not }relate to people, you may skip the remaining questions in this section.

\begin{itemize}

\item \textbf{Does the dataset identify any subpopulations (e.g., by
    age, gender)?} If so, please describe how these subpopulations are
  identified and provide a description of their respective
  distributions within the dataset.
\\
\answer{FLAIR data is only annotated with the Flickr user id and does not explicitly identify any traits.}

\item \textbf{Is it possible to identify individuals (i.e., one or
    more natural persons), either directly or indirectly (i.e., in
    combination with other data) from the dataset?} If so, please
    describe how.
\\
\answer{No. Images with personal identifiable information have been removed from FLAIR.}

\item \textbf{Does the dataset contain data that might be considered
    sensitive in any way (e.g., data that reveals rac{e} or ethnic
    origins, sexual orientations, religious beliefs, political
    opinions or union memberships, or locations; financial or health
    data; biometric or genetic data; forms of government
    identification, such as social security numbers; criminal
    history)?} If so, please provide a description.
\\
\answer{No.}


\end{itemize}

\subsection{Collection Process}
\label{app:collect}
As with the {questions in the} previous section, dataset creators should
read through these questions prior to any data collection to flag
potential issues and then provide answers once collection is complete.
{In addition to the goals outlined in the previous section, the
questions in this section are designed to elicit information that may
help researchers and practitioners to create alternative datasets with
similar characteristics. Again, questions that apply only to datasets
that relate to people are grouped together at the end of the
section.}

\begin{itemize}

\item \textbf{How was the data associated with each instance
    acquired?} Was the data directly observable (e.g., raw text, movie
  ratings), reported by subjects (e.g., survey responses), or
  indirectly inferred/derived from other data (e.g., part-of-speech
  tags, model-based guesses for age or language)? If {the} data was reported
  by subjects or indirectly inferred/derived from other data, was the
  data validated/verified? If so, please describe how.
\\
\answer{Images and associated image ID and user ID were acquired from the Flickr website. The data was directly observable.}

\item \textbf{What mechanisms or procedures were used to collect the
    data (e.g., hardware apparatus{es} or sensor{s}, manual human
    curation, software program{s}, software API{s})?} How were these
    mechanisms or procedures validated?
\\
\answer{Software API provided by Flickr.}

\item \textbf{If the dataset is a sample from a larger set, what was
    the sampling strategy (e.g., deterministic, probabilistic with
    specific sampling probabilities)?}
\\
\answer{N/A.}

\item \textbf{Who was involved in the data collection process (e.g.,
    students, crowdworkers, contractors) and how were they compensated
    (e.g., how much were crowdworkers paid)?}
\\
\answer{N/A.}

\item \textbf{Over what timeframe was the data collected?} Does this
  timeframe match the creation timeframe of the data associated with
  the instances (e.g., recent crawl of old news articles)?  If not,
  please describe the timeframe in which the data associated with the
  instances was created.
\\
\answer{The images were collected from late 2017 to early 2018.}

\item \textbf{Were any ethical review processes conducted (e.g., by an
    institutional review board)?} If so, please provide a description
  of these review processes, including the outcomes, as well as a link
  or other access point to any supporting documentation.
\\
\answer{No.}

\end{itemize}

{If the dataset does not relate to people, you may skip the remaining questions in this section.}

\begin{itemize}

\item \textbf{Did you collect the data from the individuals in
    question directly, or obtain it via third parties or other sources
    (e.g., websites)?}
\\
\answer{The data was collected from the Flickr website.}

\item \textbf{Were the individuals in question notified about the data
    collection?} If so, please describe (or show with screenshots or
  other information) how notice was provided, and provide a link or
  other access point to, or otherwise reproduce, the exact language of
  the notification itself.
\\
\answer{N/A.} 

\item \textbf{Did the individuals in question consent to the
    collection and use of their data?} If so, please describe (or show
  with screenshots or other information) how consent was requested and
  provided, and provide a link or other access point to, or otherwise
  reproduce, the exact language to which the individuals consented.
\\
\answer{Each image has an associated license chosen by the Flickr user. 
FLAIR only contain images with one of the following permissive licenses:
\begin{itemize}
    \item Attribution 2.0 Generic (CC BY 2.0)\footnote{\url{https://creativecommons.org/licenses/by/2.0/}}
    \item Attribution-ShareAlike 2.0 Generic (CC BY-SA 2.0) \footnote{\url{https://creativecommons.org/licenses/by-sa/2.0/}}
    \item Attribution-NoDerivs 2.0 Generic (CC BY-ND 2.0)\footnote{\url{https://creativecommons.org/licenses/by-nd/2.0/}}
    \item U.S. Government Works \footnote{\url{http://www.usa.gov/copyright.shtml}}
    \item CC0 1.0 Universal (CC0 1.0)
Public Domain Dedication \footnote{\url{https://creativecommons.org/publicdomain/zero/1.0/}}
    \item Public Domain Mark 1.0 \footnote{\url{https://creativecommons.org/publicdomain/mark/1.0/}}
\end{itemize}
}

\item \textbf{If consent was obtained, were the consenting individuals
    provided with a mechanism to revoke their consent in the future or
    for certain uses?} If so, please provide a description, as well as
  a link or other access point to the mechanism (if appropriate).
\\
\answer{N/A.}

\item \textbf{Has an analysis of the potential impact of the dataset
    and its use on data subjects (e.g., a data protection impact
    analysis) been conducted?} If so, please provide a description of
  this analysis, including the outcomes, as well as a link or other
  access point to any supporting documentation.
\\
\answer{N/A.}
\item \textbf{Any other comments?}
\\
\answer{
After initial collection, we applied a two-stage filtering approach to remove images with personal identifiable information and sensitive materials.
In the first stage, we used a face detector to automatically remove images with faces. 
In the second stage, we asked human annotators to filter out images with identifiable human and sensitive materials.
Specifically, images with any of the following will be removed from FLAIR:
\begin{itemize}
    \item Visible faces or part of visible faces.
    \item Visible facial features or part of visible facial features, such as hair, eye, eyebrow, mouth, nose, ear, etc.
    \item Human body or part of body has identifiable feature, such as tatto, disabilities, injuries, scars, birthmarks, unique moles, etc.
    \item Rude statements and expressions.
    \item Profanity, racial, gender, ethnic, or religious slurs.
    \item Sexually explicit or pornographic materials.
    \item Violent, obscene, graphic or disturbing materials.
\end{itemize}
}
\end{itemize}

\subsection{Preprocessing/cleaning/labeling}

Dataset creators should read through these questions prior to any
preprocessing, cleaning, or labeling and then provide answers once
these tasks are complete. The questions in this section are intended
to provide dataset consumers with the information they need to
determine whether the ``raw'' data has been processed in ways that are
compatible with their chosen tasks. For example, text that has been
converted into a ``bag-of-words'' is not suitable for tasks involving
word order.

\begin{itemize}

\item \textbf{Was any preprocessing/cleaning/labeling of the data done
    (e.g., discretization or bucketing, tokenization, part-of-speech
    tagging, SIFT feature extraction, removal of instances, processing
    of missing values)?} If so, please provide a description. If not,
  you may skip the remain{ing} questions in this section.
\\
\answer{Yes.
Labeling was done by human annotators where one annotator labeled the objects presented in an image and another annotator validate the labeling.
The taxonomy of the labels were constructed as following:
\begin{enumerate}
    \item Retrieve all keywords from ShutterStock~\footnote{\url{https://www.shutterstock.com/}} attached to 1000 images or more.
    \item Remove keywords that are illicit substances, sexual content, negative connotations, adjectives, proper names, places, organizations, occupations, abstract concepts, references to ethnicity, culture, religion, skin color, all body parts, and most animal parts.
    \item Remove plurals, alternative spellings and synonyms.
    \item Leverage WordNet~\footnote{\url{https://wordnet.princeton.edu/}} to construct coarse-grained labels.
\end{enumerate}
Unqualified images are removed as described in Appendix~\ref{app:collect}.
}

\item \textbf{Was the ``raw'' data saved in addition to the preprocessed/cleaned/labeled data (e.g., to support unanticipated future uses)?} If so, please provide a link or other access point to the ``raw'' data.
\\
\answer{No.} 

\item \textbf{Is the software {that was} used to preprocess/clean/label the {data} available?} If so, please provide a link or other access point.
\\
\answer{The script to process data for training is provided at \url{https://github.com/apple/ml-flair}.
}


\end{itemize}

\subsection{Uses}

{The} questions {in this section} are intended to encourage dataset
creators to reflect on the tasks for which the dataset should and
should not be used. By explicitly highlighting these tasks, dataset
creators can help dataset consumers to make informed decisions,
thereby avoiding potential risks or harms.

\begin{itemize}

\item \textbf{Has the dataset been used for any tasks already?} If so, please provide a description.
\\
\answer{FLAIR has been used to benchmark federated learning and differential privacy on multi-label classification task, in this current paper.}

\item \textbf{Is there a repository that links to any or all papers or systems that use the dataset?} If so, please provide a link or other access point.
\\
\answer{The current paper and the code used for experiments are available at \url{https://github.com/apple/ml-flair}}

\item \textbf{What (other) tasks could the dataset be used for?}
\\
\answer{FLAIR could be used for other image classification tasks.}

\item \textbf{Is there anything about the composition of the dataset or the way it was collected and preprocessed/cleaned/labeled that might impact future uses?} For example, is there anything that a {dataset consumer} might need to know to avoid uses that could result in unfair treatment of individuals or groups (e.g., stereotyping, quality of service issues) or other {risks or} harms (e.g., {legal risks,} financial harms{)?} If so, please provide a description. Is there anything a {dataset consumer} could do to mitigate these {risks or} harms?
\\
\answer{This dataset contains a limited number of object classes and is intended to create a benchmark to evaluate and compare algorithms for (private) federated learning. }

\item \textbf{Are there tasks for which the dataset should not be used?} If so, please provide a description.
\\
\answer{It being a subset of images from Flickr, it is not expected to be representative of all images in the world.}


\end{itemize}

\subsection{Distribution}

Dataset creators should provide answers to these questions prior to
distributing the dataset either internally within the entity on behalf
of which the dataset was created or externally to third parties.

\begin{itemize}

\item \textbf{Will the dataset be distributed to third parties outside of the entity (e.g., company, institution, organization) on behalf of which the dataset was created?} If so, please provide a description.
\\
\answer{Yes.}

\item \textbf{How will the dataset will be distributed (e.g., tarball on website, API, GitHub)?} Does the dataset have a digital object identifier (DOI)?
\\
\answer{The dataset will be distributed on AWS S3.}

\item \textbf{When will the dataset be distributed?}
\\
\answer{The dataset will be distributed on June 16th, 2022.}

\item \textbf{Will the dataset be distributed under a copyright or other intellectual property (IP) license, and/or under applicable terms of use (ToU)?} If so, please describe this license and/or ToU, and provide a link or other access point to, or otherwise reproduce, any relevant licensing terms or ToU, as well as any fees associated with these restrictions.
\\
\answer{Please see license for FLAIR at \url{https://github.com/apple/ml-flair/blob/master/LICENSE.md}}

\item \textbf{Have any third parties imposed IP-based or other restrictions on the data associated with the instances?} If so, please describe these restrictions, and provide a link or other access point to, or otherwise reproduce, any relevant licensing terms, as well as any fees associated with these restrictions.
\\
\answer{No.}

\item \textbf{Do any export controls or other regulatory restrictions apply to the dataset or to individual instances?} If so, please describe these restrictions, and provide a link or other access point to, or otherwise reproduce, any supporting documentation.
\\
\answer{N/A.}


\end{itemize}

\subsection{Maintenance}

As with the {questions in the} previous section, dataset creators
should provide answers to these questions prior to distributing the
dataset. The questions {in this section} are intended to
encourage dataset creators to plan for dataset maintenance and
communicate this plan {to} dataset consumers.

\begin{itemize}

\item \textbf{Who {will be} supporting/hosting/maintaining the dataset?}
\\
\answer{Apple ML Privacy team.}

\item \textbf{How can the owner/curator/manager of the dataset be contacted (e.g., email address)?}
\\
\answer{\url{pfl-dev@group.apple.com}}

\item \textbf{Is there an erratum?} If so, please provide a link or other access point.
\\
\answer{N/A.}

\item \textbf{Will the dataset be updated (e.g., to correct labeling
    errors, add new instances, delete instances)?} If so, please
  describe how often, by whom, and how updates will be communicated to
  {dataset consumers} (e.g., mailing list, GitHub)?
\\
\answer{N/A.}

\item \textbf{If the dataset relates to people, are there applicable
    limits on the retention of the data associated with the instances
    (e.g., were {the} individuals in question told that their data would
    {be} retained for a fixed period of time and then deleted)?} If so,
    please describe these limits and explain how they will be
    enforced.
\\
\answer{N/A.}

\item \textbf{Will older versions of the dataset continue to be
    supported/hosted/maintained?} If so, please describe how. If not,
  please describe how its obsolescence will be communicated to {dataset
  consumers}.
\\
\answer{N/A.}

\item \textbf{If others want to extend/augment/build on/contribute to
    the dataset, is there a mechanism for them to do so?} If so,
  please provide a description. Will these contributions be
  validated/verified? If so, please describe how. If not, why not? Is
  there a process for communicating/distributing these contributions
  to {dataset consumers}? If so, please provide a description.
\\
\answer{N/A.}

\item \textbf{Any other comments?}
\\
\answer{The annotations and Apple’s other rights in the dataset are licensed under CC-BY-NC 4.0 license. The images are copyright of the respective owners, the license terms of which can be found using the links provided in \url{https://github.com/apple/ml-flair/blob/master/ATTRIBUTION.txt} (by matching the Image ID). Apple makes no representations or warranties regarding the license status of each image and you should verify the license for each image yourself.}

\end{itemize}

\section{Benchmark Setup Details}
\subsection{Computational resources}
All experiments are conducted on a cluster with 32 CPU cores and 4 NVIDIA Tesla V100 GPUs.

\subsection{Hyper-parameters grids}
Below are the hyper-parameter grids that we searched on for benchmarking FLAIR:
\label{app:hyper}
\begin{itemize}
    \item Server learning rate $\in\{0.01, 0.02, 0.05, 0.1\}$
    \item Server number of rounds $\in\{2000, 5000\}$
    \item Client local learning rate $\in\{0.01, 0.1\}$
    \item Client number of epochs $\in\{1, 2, 3, 4, 5\}$
    \item Target unclipped quantile for adaptive clipping $\in\{0.1, 0.2\}$
\end{itemize}

\subsection{Additional binary classification benchmark}

\begin{table}[h]
\centering
\caption{FLAIR binary classification benchmark results on test set for \emph{structure} label. AP stands for averaged precision.}
\footnotesize
\begin{tabular}{ll|rrrr}
\toprule
Setting & Initialization & AP & Precision & Recall & F1\\
\midrule
Centralized & Random & 87.72 & 79.76 & 78.38 & 79.06  \\
Federated  & Random & 84.22 & 77.18 & 73.42 & 75.25  \\
Private federated & Random & 68.56 & 64.33 & 76.08 & 69.71 \\
\midrule
Centralized & ImageNet & 92.80 & 84.58 & 83.99 & 84.28  \\
Federated & ImageNet & 90.49 & 81.98 & 81.65 & 81.81  \\
Private federated & ImageNet & 83.41 & 76.95 & 71.90 & 74.34 \\
\bottomrule
\end{tabular}
\label{tab:binary}
\end{table}

We provide additional binary classification benchmark on the most common \emph{structure} label, using the same hyperparameters as in Section~\ref{sec:setup}.
Table~\ref{tab:binary} summarizes the results.
The performance of models trained in federated setting and private federated setting are much closer to the centralized setting, especially when the models were pretrained on ImageNet.
We believe that this simple binary classification baseline could help researchers to quickly verify their proposed algorithms and methods in (private) federated learning setting.